\documentclass[letterpaper, 10 pt, conference]{ieeeconf}  
\IEEEoverridecommandlockouts 
\usepackage{amsfonts}
\usepackage{amsmath}
\usepackage{amssymb}
\usepackage{amsthm}
\usepackage{array}
\usepackage{epsfig}
\usepackage{epstopdf}
\usepackage{graphicx}
\usepackage{placeins}
\usepackage{subfigure}
\usepackage{wrapfig}
\usepackage{url}

\theoremstyle{definition}

\newtheorem{prob}{Problem}[section]

\graphicspath{{figs/}}

\title{\LARGE \bf Viewpoint Selection for Grasp Detection}

\author{Marcus Gualtieri and Robert Platt$^*$
\thanks{$^*$College of Computer and Information Science, Northeastern University, Boston, Massachusetts, USA,  {\tt\small mgualti@ccs.neu.edu}. This work has been supported in part by the National Science Foundation through IIS-1427081, NASA through NNX16AC48A and NNX13AQ85G, and ONR through N000141410047.}
}

\begin{document}

\maketitle

\thispagestyle{empty}
\pagestyle{empty}

\begin{abstract}

In grasp detection, the robot estimates the position and orientation of potential grasp configurations directly from sensor data. This paper explores the relationship between viewpoint and grasp detection performance. Specifically, we consider the scenario where the approximate position and orientation of a desired grasp is known in advance and we want to select a viewpoint that will enable a grasp detection algorithm to localize it more precisely and with higher confidence. Our main findings are that the right viewpoint can dramatically increase the number of detected grasps and the classification accuracy of the top-$n$ detections. We use this insight to create a viewpoint selection algorithm and compare it against a random viewpoint selection strategy and a strategy that views the desired grasp head-on. We find that the head-on strategy and our proposed viewpoint selection strategy can improve grasp success rates on a real robot by 8\% and 4\%, respectively. Moreover, we find that the combination of the two methods can improve grasp success rates by as much as 12\%.

\end{abstract}

\section{Introduction}
\label{sec:introduction}

Grasp detection has become an important framework for perception for grasping \cite{Saxena2008,Lenz2015,Redmon2015,Fischinger2013,Detry2013,Herzog2012,Kappler2015,tenPas2015,Gualtieri2016,Johns2016,Kanoulas2016}. We define \textit{grasp detection} as any system that localizes grasps directly in visual sensory data without reference to prior geometric models of objects. Figure~\ref{fig:gpdExample} (c) shows an example. Using only the visible point cloud and without a model of the object to be grasped, the system locates hand configurations from which the object can be grasped (shown as yellow grippers in the figure). Each gripper denotes the 6-DOF pose of a robotic hand from which a grasp would be feasible if the gripper fingers were to close. Relative to more traditional approaches to perception for grasping based on estimating the exact pose of the object to be grasped, grasp detection can be very effective for novel objects in cluttered environments. Although grasp detection works well in general, our prior work suggests the quality and amount of sensor data available could have a significant impact on performance \cite{Gualtieri2016}. In particular, when given relatively complete point cloud data generated using InfiniTAM, a metric SLAM software \cite{Kahler2015}, grasp detection was able to produce a 93\% grasp success rate for novel object grasping in dense clutter. However, grasp success rates dropped to 84\% for the same algorithm and the same robot when using data produced using two, fixed depth sensors. This inspired us to look at the relationship between viewpoint and the performance of a given grasp detection algorithm.

\begin{figure}
  \centering
  \vspace{0.1in}
  \subfigure[]{\includegraphics[height=1.3in, trim={3in 1in 3in 0.0in}, clip]{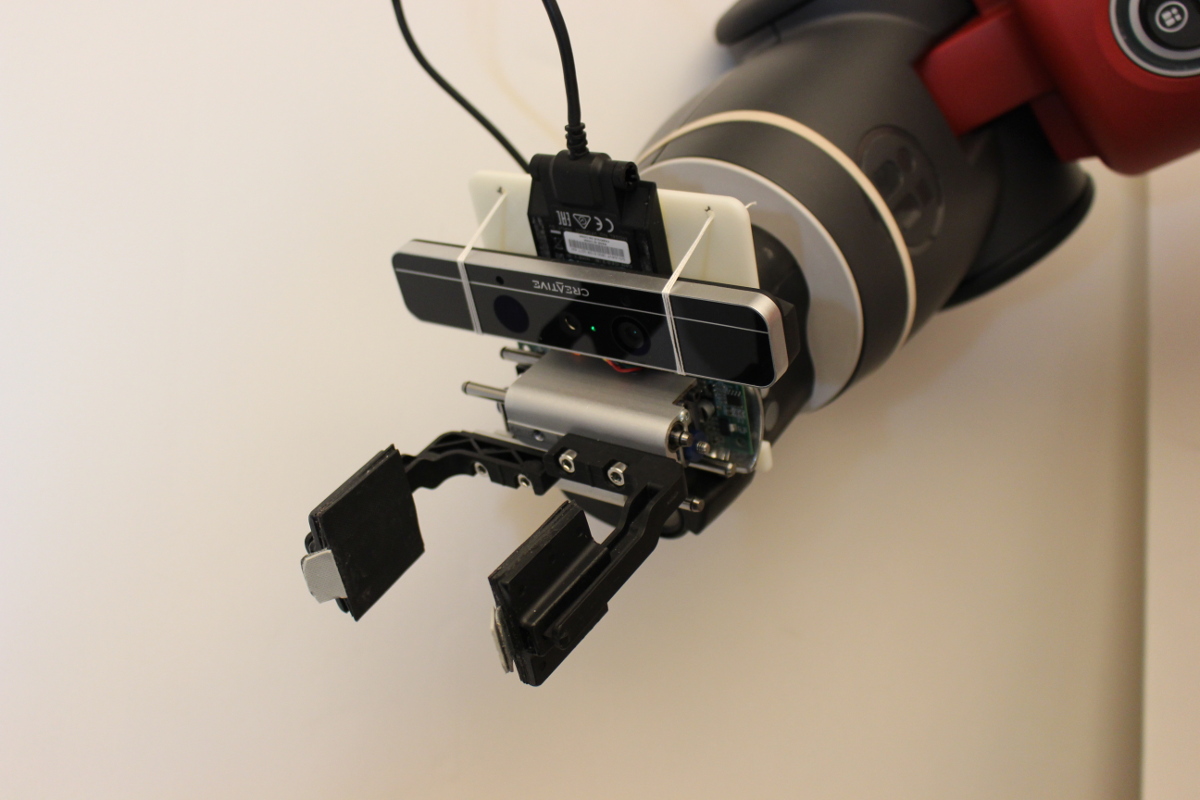}}
  \subfigure[]{\includegraphics[height=1.3in]{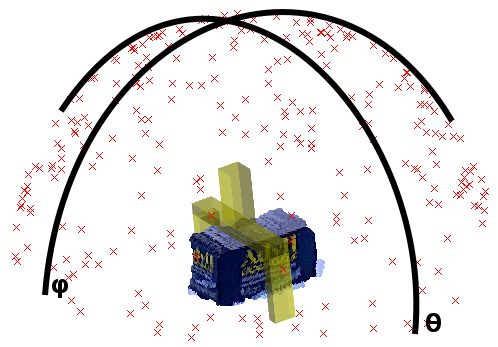}}
  \caption{(a) The eye-in-hand configuration used for our robot experiments. (b) The viewpoint selection problem. From which viewpoint should this object be viewed in order to detect a grasp nearby the one shown?}
  \vspace{-0.15in}
\label{fig:wristSensor}
\end{figure}

In this paper, we ask whether it is possible to affect the accuracy of grasp detection by selecting the right viewpoint. Specifically, we consider the scenario where the approximate position and orientation (pose) of the desired grasp is known in advance and we want to select a viewpoint in order to localize it more accurately and with higher confidence. For example, imagine the robot has already taken an RGBD image of the environment from an arbitrary viewpoint and identified an object of interest. The robot \textit{thinks} it knows how it wants to grasp the object, but it would like to take a second look to improve its confidence. Figure~\ref{fig:wristSensor} (b) illustrates this scenario. The robot thinks the grasp is located as shown in the image, but it would like to select a viewpoint from which to take a second look and ``confirm'' the detection.

Our main finding is that viewpoint most significantly affects the number of grasps detected. With the right viewpoint, the system can increase the number of detected grasps by a factor of 6 (Table~\ref{table:viewpointSelectionResults}). This is important because a larger number of grasp detections can give the robot more choice about exactly which grasp it chooses to execute. Also, it enables the robot to be more selective about which grasps it accepts, thereby effectively increasing the accuracy for the most highly ranked $n$ grasp detections. Based on these findings, we propose a ``smart'' viewpoint selection algorithm and compare it offline against random viewpoint selection and a baseline that always selects a head-on view. These experiments show that the smart viewpoint selection strategy outperforms the random strategy and that it can also outperform the head-on strategy for some classes of objects. We also evaluate the method in terms of grasp success rates in a real robotic system. Here, we used the head-on view only to ``align'' the desired grasp with the point cloud. We find that the smart viewpoint outperforms random by an average of 4\% while the head-on alignment view outperforms random by 8\%. By combining both methods (choosing the smart viewpoint, detecting grasps, and then choosing a head-on view for the grasp detected from the smart viewpoint), we outperform the random baseline by an average of 12\%.

\section{Background and Related Work}
\label{sec:backgroundAndRelatedWork}

\subsection{Active vision}
\label{sec:activeVision}

This work falls broadly into the category of research called ``active vision'', which is defined to be any scenario where the robot applies a strategy for sensor placement/configuration to perform its task \cite{Chen2011}. Chen \textit{et al.} provide a broad (although now somewhat dated) survey of the subject \cite{Chen2011}. Our paper is more specifically related to the idea of planning sensor placements for the task of object recognition. Roy \textit{et al.} survey works that benefit object recognition by purposive sensor placements \cite{Roy2004}, and Velez \textit{et al.} plan viewing trajectories to improve the performance of an off-the-shelf object detector \cite{Velez2011}. In contrast, our task is not to recognize object instances but instead to recognize grasps that are likely to succeed when executed; although, we expect some of the ideas could extend to a more general class of detection problems.

An issue with any viewpoint selection method is deciding a metric to use for viewpoint quality. Chen \textit{et al.} suggest that the best metric to use is likely to be task-dependent \cite{Chen2011}. On the other hand, general trends can be observed in the literature. One idea is to try to increase the amount of information of the scene by maximizing over Shannon entropy \cite{Vazquez2001, Kriegel2015}, KL divergence \cite{Hoof2012}, or Fisher information \cite{Levine2010}. These methods all require specifying a probability distribution over which to compute the information metric. In contrast, we directly evaluate the performance of the grasp detection system from various viewpoints and store the result in a database, which could be viewed as a non-parametric, nearest-neighbor approach \cite{Bentley1975}. We argue that this approach is simpler to implement and directly applicable to the task at hand.

In this work we restrict our attention to single views; although, constructing a scene from multiple views can sometimes be a powerful approach. We and others have explored this with significant benefit to grasp detection performance \cite{Kahn2015, Gualtieri2016}, but the primary difficulty faced with this is that the ICP-based SLAM algorithms do not register multiple views well in near-field, uncluttered scenes.

\subsection{Grasp detection}
\label{sec:graspDetection}

\begin{figure}
  \centering
  \vspace{0.1in}
  \subfigure[]{\includegraphics[height=1.4in]{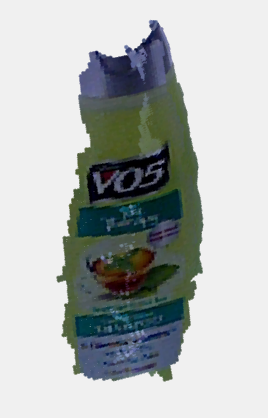}}
  \subfigure[]{\includegraphics[height=1.4in]{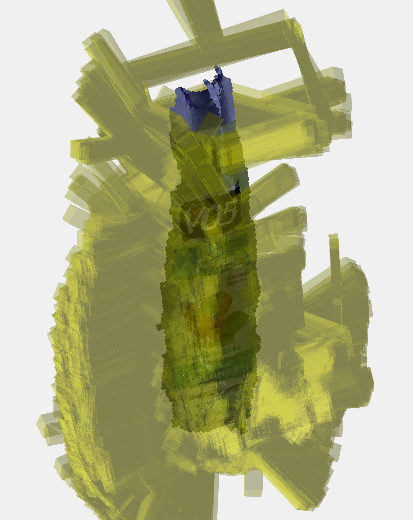}}
  \subfigure[]{\includegraphics[height=1.4in]{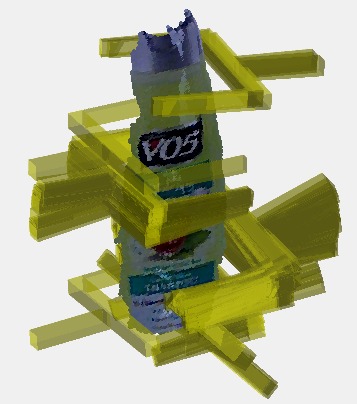}}
  \caption{(a) Input point cloud; (b) grasp candidates that denote potential grasp configurations; (c) high scoring grasps.}
  \vspace{-0.15in}
\label{fig:gpdExample}
\end{figure}

In order to understand the problem of viewpoint selection for grasp detection, it is important to understand grasp detection itself. In this paper, we use the grasp detector described in our prior work \cite{Gualtieri2016} (and, for comparison, a modification of that detector designed to be similar to the detector proposed by~\cite{Kappler2015}). This system takes point clouds as input and produces predicted grasp poses in $\textit{SE}(3)$~\footnote{$\textit{SE}(3)$ is the 6-DOF space of rigid body transformations.} as output. Each grasp pose output by the system is predicted to be a force closure grasp (i.e. a \textit{positive} grasp) for a two-fingered gripper.~\footnote{In fact, it is possible to use any grasp quality metric without changing the algorithm at all. The important thing is that there is a consistent, well-defined definition of a grasp.}

Our grasp detection algorithm proceeds in two main steps. First, given an input point cloud (Figure~\ref{fig:gpdExample} (a)), a set of 6-DOF grasp candidates are sampled using basic geometric constraints (Figure~\ref{fig:gpdExample} (b)). These grasp candidates constitute a proposal distribution analogous to the object detection context. Second, a trained classifier predicts a binary grasp/not grasp label for each of these samples and assigns a probability with this prediction (Figure~\ref{fig:gpdExample} (c)). We refer to this probability as the classifier's \textit{confidence score}.

The classifier that predicts grasps needs to be trained in a supervised fashion with a labeled dataset. The ground truth labels in the dataset say which grasps are \textit{true positives} and which are \textit{true negatives}.  \textit{Accuracy}, which is the number of true positives and true negatives divided by the total number of predictions, quantifies the performance of the classifier. The labels can be generated using any reasonable grasp metric, human annotation, simulation (as in \cite{Johns2016}), or the robot itself (as in \cite{Levine2016}). In our system, we generate training data using BigBIRD~\cite{Singh2014}, a dataset comprised of 125 3D object mesh models paired with 600 RGBD images taken from a hemisphere of different perspectives around the object. From the images we generate a large number of grasp candidates that serve as training examples. The ground truth for each candidate is calculated by using the 3D mesh model to evaluate whether a force closure grasp would result if the fingers were to close in that configuration.

\section{Problem Statement}
\label{sec:problemStatement}


We assume the robot already knows approximately how it would like to pick up the object. For example, suppose the robot has detected a coffee mug that is to be picked up. The robot knows approximately how it should grasp the mug, but we would like to detect true positive grasps nearby the desired grasp with high confidence. More precisely,

\begin{prob}[Viewpoint Selection for Grasp Detection]
\textbf{Given} 1) a region of desired grasps (the center of which is called the \textit{target grasp}), 2) the geometric category of the object in the vicinity of the grasp region (e.g. box-like, cylinder-like), 3) a desired number of grasps to detect, $n$, and 4) a set of potential viewpoints; \textbf{Calculate} the viewpoint from which the target grasp should be viewed in order to maximize the average accuracy over the top $n$ scoring grasps.
\label{prob:problemStatement}
\end{prob}

In this paper, we assume the \textit{region of desired grasps} is expressed in terms of a single target grasp pose and a neighborhood of positions and orientations about this desired pose. The objective is to maximize accuracy over the top $n$ scoring grasps because, on a real robot, we typically need several grasps to choose from in order to satisfy kinematic and task constraints.

\section{Mapping Detection Performance as a Function of Viewpoint}
\label{sec:mapping}

In order to create an algorithm for selecting viewpoint, our first step is to characterize variables of interest (such as density of detected grasps, accuracy, etc.) as a function of viewpoint. Specifically, we create maps of the relevant variables where directions are azimuth ($\theta$) and elevation ($\phi$) of viewpoint relative to the target grasp. The map is created entirely using data from the BigBIRD dataset~\cite{Singh2014}, comprised of real RGBD data for a variety of objects taken under controlled conditions. Each object in BigBIRD is associated with 600 different point clouds taken from different views in a hemisphere around the object.

\subsection{Projecting viewpoint into the reference frame of a grasp}
\label{sec:projecting}

\begin{figure}
  \centering
  \vspace{0.1in}
  \includegraphics[height=1.4in]{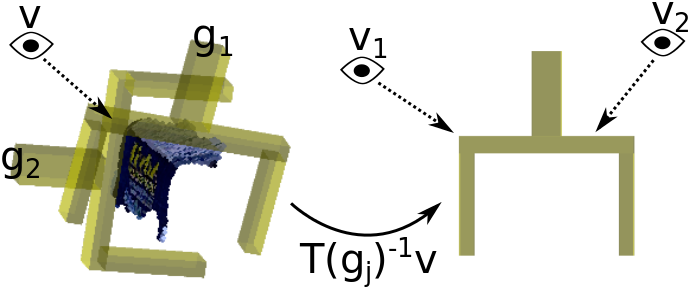}
  \caption{Illustration of viewpoint projection into the reference frames of the grasps.}
  \label{fig:viewCalculation}
  \vspace{-0.15in}
\end{figure}

Since our maps will be expressed relative to the pose of the target grasp, we need to transform viewpoint into the grasp reference frame. We do this as follows. First, we run grasp detection on point clouds taken from a set of $n$ different viewpoints $v_1, \dots, v_n \in \mathbb{R}^3$, expressed as positions in a world reference frame (we assume that the camera is always pointed directly toward the target grasp). For each viewpoint $i \in (1,n)$, let $g^i_1, \dots, g^i_{m_i}$ denote the set of $m_i$ grasps detected from that viewpoint. Then, project each viewpoint into the reference frames of the grasps seen from that viewpoint. Figure~\ref{fig:viewCalculation} illustrates this idea. The left side shows two grasps that are detected in a single point cloud generated by viewing the object from $v$. When projected into the reference frames of the two grasps, we get the two viewpoints seen on the right side. Let $T(g)$ denote the $4\times4$ homogeneous transform that expresses the reference frame of grasp $g$ relative to the world reference frame. Then, viewpoint $i$ expressed in the reference frame of grasp $j$ is $v^i_j = T(g^i_j)^{-1} v_i$.

\subsection{Creating the maps}
\label{sec:creating}

We create a map as follows. First, we detect grasps from many different viewpoints around several different objects. For each object, we sample 80 different viewpoints randomly from the 600 total views per object in BigBIRD. Then, we perform the viewpoint projection described above onto the grasps. The result is a set of viewpoints, $\mathbb{V} = (v^1_1, \dots, v^n_{m_n})$, expressed in the reference frame of the detected grasp. For each viewpoint, $v \in \mathbb{V}$, we store the confidence score assigned by the grasp detector to the corresponding grasp, $s(v) \in \mathbb{R}$, and the ground truth evaluation of whether the grasp was actually a force closure grasp or not, $gt(v) \in \{0,1\}$.

\begin{figure*}
  \centering
  \vspace{0.1in}
  \subfigure[]{\includegraphics[height=1.0in]{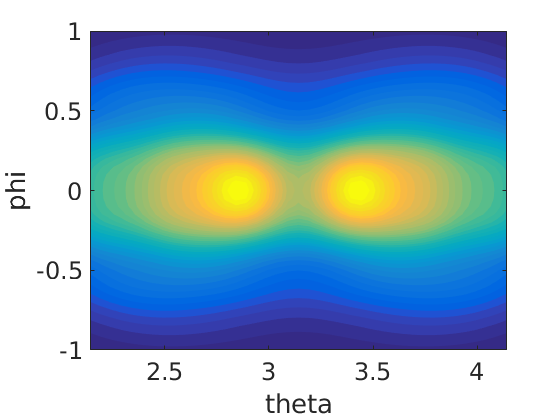}}
  \subfigure[]{\includegraphics[height=1.0in]{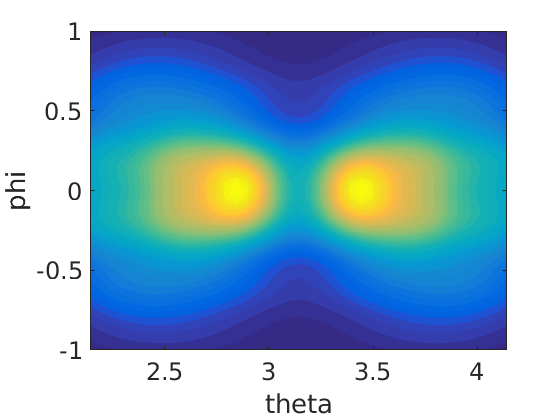}}
  \subfigure[]{\includegraphics[height=1.0in]{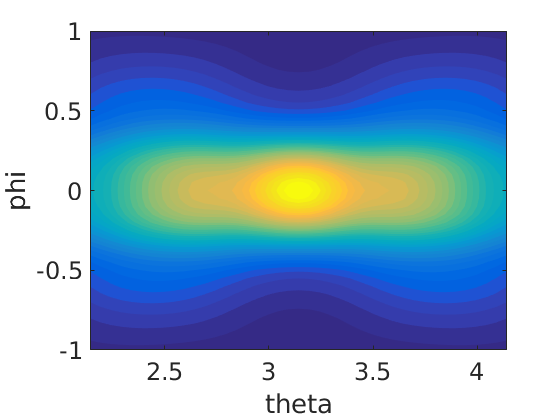}}
  \subfigure[]{\includegraphics[height=1.0in]{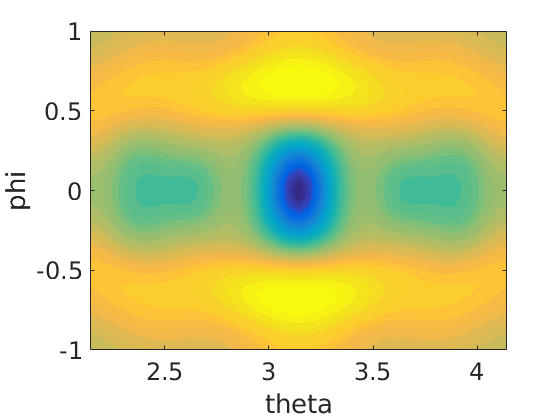}}
  \subfigure[]{\includegraphics[height=1.0in]{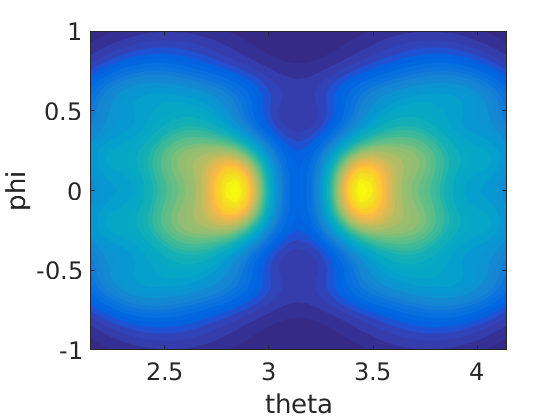}}
  \subfigure[]{\includegraphics[height=1.0in]{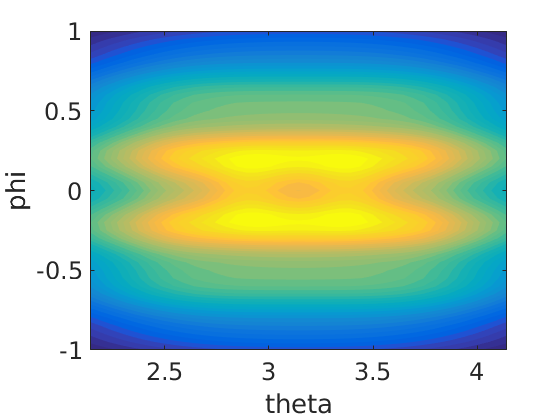}}
  \subfigure[]{\includegraphics[height=1.0in]{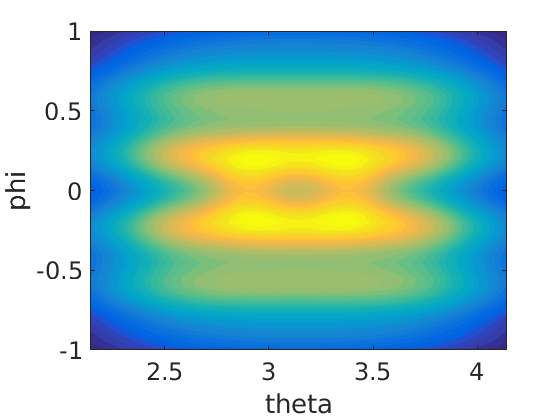}}
  \subfigure[]{\includegraphics[height=1.0in]{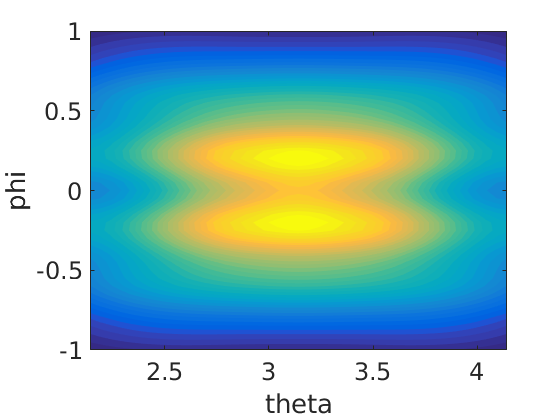}}
  \subfigure[]{\includegraphics[height=1.0in]{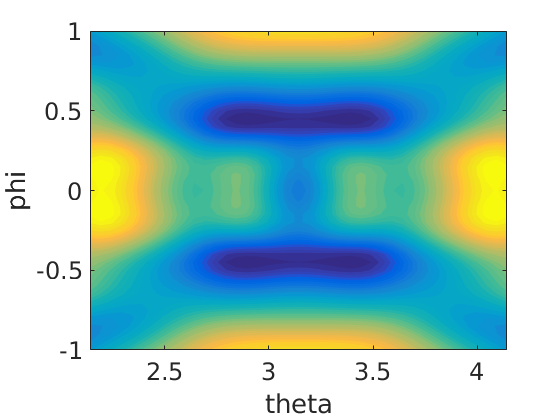}}
  \subfigure[]{\includegraphics[height=1.0in]{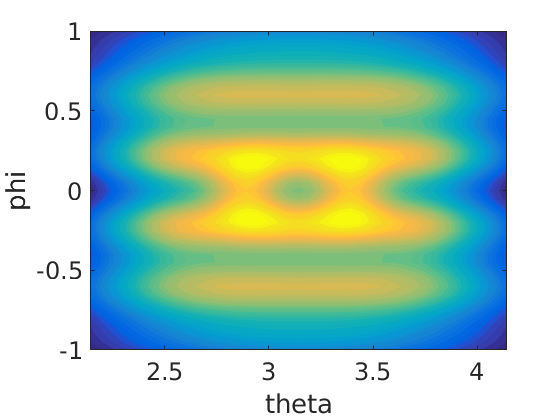}}
  \subfigure[]{\includegraphics[height=1.0in]{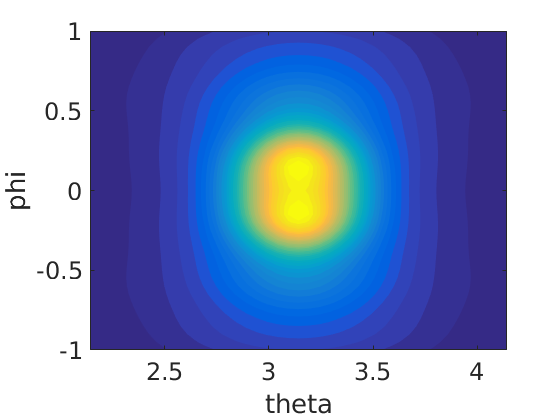}}
  \subfigure[]{\includegraphics[height=1.0in]{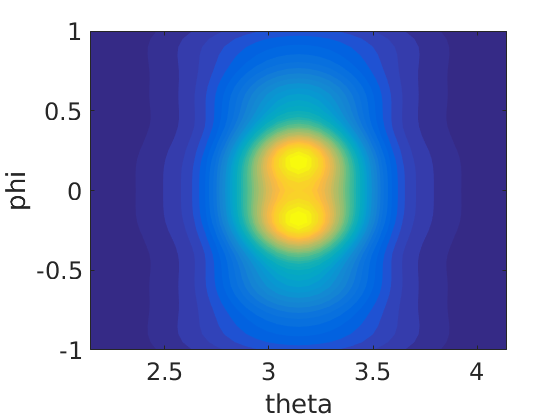}}
  \subfigure[]{\includegraphics[height=1.0in]{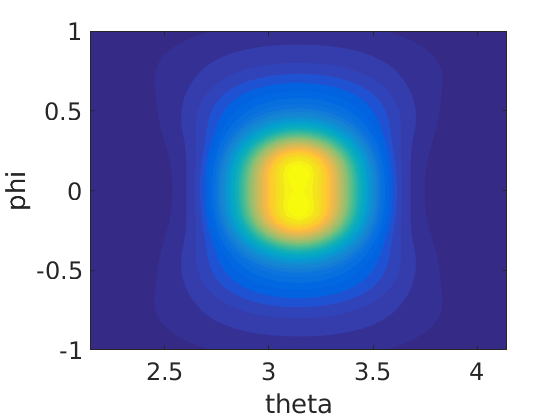}}
  \subfigure[]{\includegraphics[height=1.0in]{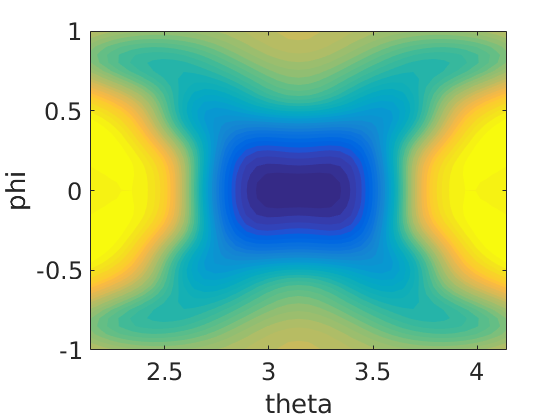}}
  \subfigure[]{\includegraphics[height=1.0in]{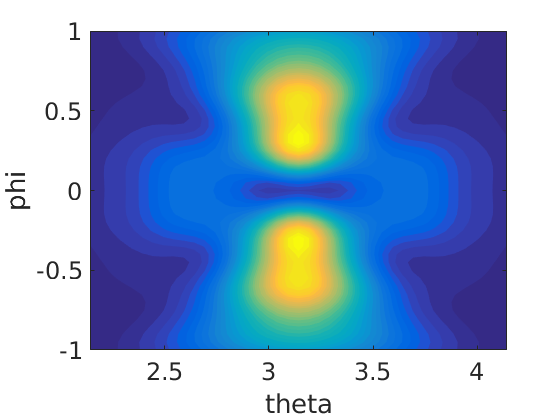}}
  \caption{Various performance metrics plotted as a function of viewpoint (azimuth and elevation). \textit{Rows:} Top: results averaged over 25 box-like objects using default algorithm. Middle: results averaged over 14 cylinder-like objects using default algorithm. Bottom: results averaged over 25 box-like objects using Kappler algorithm variant~\cite{Kappler2015}. \textit{Columns:} First: density of all considered grasp candidates. Second: density of true positives. Third: density of false positives. Fourth: accuracy. Fifth: density of true positives minus density of false positives (our proposed measure).}
  \vspace{-0.15in}
\label{fig:mapBoxesCylinders}
\end{figure*}

\begin{wrapfigure}{r}{0.25\textwidth}
  \centering
  \includegraphics[height=1.5in]{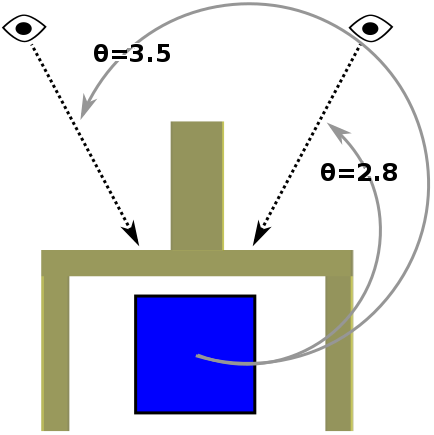}
  \caption{Elevation angle of viewpoints from which to view a target grasp on a box-like object in order to maximize number of true positive minus false positive grasps detected.}
\label{fig:bestViews}
\end{wrapfigure}


BigBIRD consists of 125 objects that can be divided into roughly two shape classes: box-like objects and cylinder-like objects. Since these two object classes appear to have qualitatively different maps, we will characterize them separately. Figure~\ref{fig:mapBoxesCylinders} (a-e) shows the results of averaging the maps created for a set of 25 box-like objects from BigBIRD. Here, viewpoint is expressed in terms of the azimuth ($\theta$) and elevation ($\phi$) of the unit vector pointing toward the viewpoint expressed in the reference frame of the grasp. (Coordinates illustrated in Figure~\ref{fig:wristSensor} (b) and Figure~\ref{fig:bestViews}.) The maps are smoothed by convolving the data with an isotropic Gaussian with variance equal to 0.2. The convolved map values are computed at discrete positions spaced 0.05 radians ($3^\circ$) and going out to $\pm 1.05$ radians ($\pm 60^\circ$) on both axes. Figure~\ref{fig:mapBoxesCylinders} (a-e) shows, respectively: the density of all detected grasps, the density of true positives (number of correct grasp detections), the density of false positives, grasp classification accuracy, and the density of true positives minus the density of false positives. Figure~\ref{fig:mapBoxesCylinders} (f-j) shows the same set of five maps, averaged over 14 cylindrical objects from BigBIRD. Notice whereas box-like objects are best viewed from a viewpoint of approximately $(\phi, \theta) = (0 \pm 0.1, 2.8 \pm 0.1)$, cylinder-like objects are best viewed from $(\phi, \theta) = (0.2 \pm 0.1, 2.9 \pm 0.2)$. Another interesting observation is that the grasp detection algorithm itself also makes a difference. The last row of Figure~\ref{fig:mapBoxesCylinders} shows the same statistics for a variant of grasp detection similar to the detector proposed by Kappler et al.~\cite{Kappler2015} for the same set of 25 box-like objects. Whereas our default method samples grasp candidates by searching orientations about a local curvature axis, the method of Kappler et al. samples grasp candidates by searching orientations about a local normal axis. For the sake of comparison, we hold the method of classification fixed. Notice the shape of the map is completely different. This suggests the grasp detection algorithm also has an impact on viewpoint. 

\section{Viewpoint Selection}
\label{sec:viewpointSelection}

\subsection{Approach}

The key question is how the maps developed in the last section should be used to guide viewpoint. Perhaps the most obvious answer is to use classification accuracy, i.e. the maps shown in Figure~\ref{fig:mapBoxesCylinders} (d, i, n). However, there are a couple of problems with this choice. First, for all three object/algorithm variations, classification accuracy seems to be maximized at viewpoints associated with low-density grasp detection. Generating a large number of detections is important because it gives us a larger number of high scoring grasps to choose from, which will be crucial for a robot with kinematic and task constraints. Second, we do not empirically observe a significant improvement in accuracy in our offline evaluations by selecting viewpoint based on accuracy. It could be the expected improvement in accuracy is not large enough to make a difference, and it could be there is wider variation among objects as to which viewpoint maximizes detection accuracy.

Instead, we propose selecting the viewpoint that maximizes the expected number of true positives detected minus the expected number of false positives. This measure is shown in the right-most column of Figure~\ref{fig:mapBoxesCylinders}. The idea is we want to maximize the number of true positives and minimize the number of false positives. An algorithm that selects viewpoints that maximizes this measure should generate a large number of true positives (i.e. a large number of correctly identified grasps) and thereby increase accuracy among the highest scoring grasps.

\subsection{Offline experiments}

We ran experiments to compare grasp detection performance using 1) viewpoints chosen using the true positives minus false positives density map (\textit{smart}), 2) viewpoints selected uniformly randomly from the set of possible viewpoints (\textit{random}), and 3) head-on views of the target grasp (\textit{head-on}). In the head-on contingency, we select from the set of possible viewpoints the one closest to a view that would see the object along the approach vector of the grasp. We performed this comparison offline using the point clouds contained within the BigBIRD dataset as follows. First, for a given object, select a ground truth grasp at random (using the mesh to evaluate ground truth). This will be the target grasp.~\footnote{In practice, the target grasp would be selected based on task or object characteristics, but here it is selected randomly.} Second, for this target grasp, select a viewpoint using the proposed method, the head-on method, or at random out of the 600 views available for each BigBIRD object. Third, run grasp detection for the point cloud obtained from that view and prune out all grasps that are not within the neighborhood of the desired grasp. We evaluate the number of detected grasps and the detection accuracy of the top $n$ scoring detections averaged over all objects within a category.

\begin{figure}
  \centering
  \vspace{0.1in}
  \subfigure[]{\includegraphics[height=1.2in]{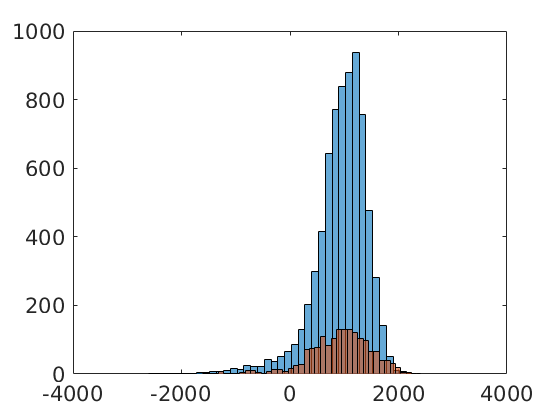}}
  \subfigure[]{\includegraphics[height=1.2in]{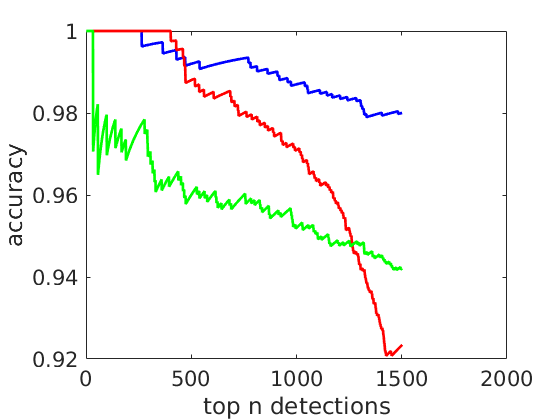}}
  \caption{Grasp detection performance averaged over 25 box-like objects in BigBIRD: (a) histogram over the scores of the detected grasps using smart viewpoints (blue) and random viewpoints (red); (b) detection accuracy averaged over the top $n$ scoring detected grasps. Blue: smart viewpoints. Green: head-on viewpoints. Red: random viewpoints.}
  \vspace{-0.15in}
\label{fig:histogram_boxes_curvature}
\end{figure}

\begin{table}[h]
  \centering
  \begin{tabular}{c | c | c | c} 
    \hline
                        & True Positives & Positives & Accuracy \\
    \hline
    Boxes Smart         & 5448  & 6123  & 0.89 \\
    \hline
    Boxes Head-on       & 1923  & 2206  & 0.87 \\
    \hline
    Boxes Random        & 1312  & 1424  & 0.92 \\
    \hline
    Cylinders Smart     & 13834 & 17150 & 0.80 \\
    \hline
    Cylinders Head-on   & 14291 & 18112 & 0.79 \\
    \hline
    Cylinders Random    & 2155  & 2754  & 0.78 \\
    \hline
  \end{tabular}
  \caption{Summary of viewpoint selection results. Counts are totaled from all viewpoints and objects, and accuracy is averaged over all viewpoints and objects.}
  \vspace{-0.15in}
  \label{table:viewpointSelectionResults}
\end{table}

Figure~\ref{fig:histogram_boxes_curvature} shows the results averaged over the 25 box-like objects in BigBIRD. Figure~\ref{fig:histogram_boxes_curvature} (a) shows a histogram over scores for all positives detected. The larger blue histogram shows the scores of the grasps detected using smart viewpoints. The smaller red histogram shows the same for random viewpoints. Figure~\ref{fig:histogram_boxes_curvature} (b) shows accuracy for the top $n$ scoring grasps using the three viewpoint selection methods. In each contingency, we detected a large number of grasps over all objects within the category (either box-like or cylinder-like) and pruned those that were not within the desired neighborhood of the target grasp (2 cm and $20^\circ$) in our experiments). Of the remainder, we ranked them by the score produced by the deep network used to classify the grasps (higher means more ``confident'') and evaluated classification accuracy over the top $n$ scoring grasps. Out of the three methods, the head-on view (the green line in Figure~\ref{fig:histogram_boxes_curvature} (b)) is clearly at a disadvantage. The reason can be seen by looking at the maps for true positives and accuracy (Figure~\ref{fig:mapBoxesCylinders} (c, d)). The head-on view ``sees'' the target grasp from $\phi=0, \theta=0$. However, this view is associated with low classification accuracy and high numbers of false positives. Also, notice that the top-$n$ accuracy for random views (the red line in Figure~\ref{fig:histogram_boxes_curvature} (b)) drops off significantly for larger values of $n$. Even though random does happen to detect some good grasps, it does not detect as many of them as does the proposed method (the blue line in Figure~\ref{fig:histogram_boxes_curvature} (b)). 

\begin{figure}
  \centering
  \vspace{0.1in}
  \subfigure[]{\includegraphics[height=1.2in]{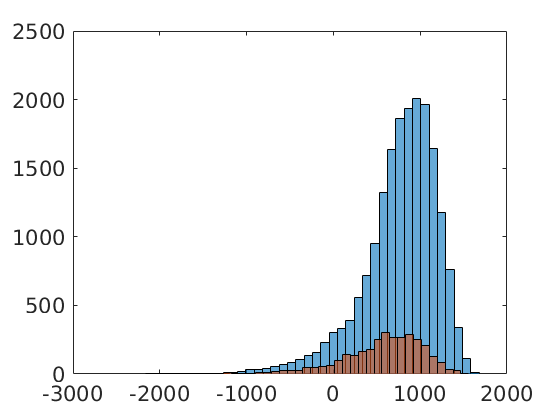}}
  \subfigure[]{\includegraphics[height=1.2in]{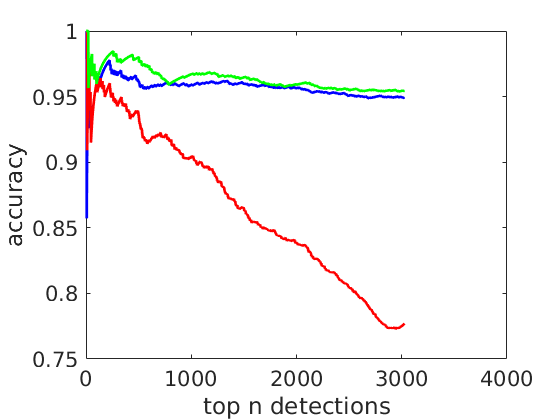}}
  \caption{Grasp detection performance averaged over 14 cylinder-like objects in BigBIRD: (a) histogram over the scores of the detected grasps using smart viewpoints (blue) and random viewpoints (red); (b) detection accuracy averaged over the top $n$ scoring detections. Blue: smart viewpoints. Green: head-on viewpoints. Red: random viewpoints.}
  \vspace{-0.15in}
  \label{fig:histogram_other}
\end{figure}

Figure~\ref{fig:histogram_other} shows similar results for the group of 14 cylindrical objects in BigBIRD. Figure~\ref{fig:histogram_other} (a) shows that the random viewpoint method detects approximately 6 times fewer positive grasps than does our proposed method. Figure~\ref{fig:histogram_other} (b) shows that top-$n$ classification accuracy for random viewpoints (the red line) drops off quickly, as it did for box-like objects. Interestingly, the head-on method performs similarly to the proposed method for cylinder-like objects. This is because the head-on view for cylinders turns out to be a relatively good view (see $\phi=0, \theta=0$ in Figure~\ref{fig:mapBoxesCylinders} (g, i)). For cylinder-like objects, the proposed method selects views that nearly approximate the head-on view.

\section{Robot Experiments}
\label{sec:robotExperiments}

The results reported so far indicate that our proposed viewpoint selection method can improve the accuracy with which the top $n$ grasps are detected. But, how well does this translate into grasp success on a real robotic system? In this section, we evaluate the approach in the context of a robot grasping in dense clutter.

\subsection{Setup}

\begin{figure}
  \centering
  \vspace{0.1in}
  \subfigure[First view (random)]{
    \includegraphics[height=1.3in,trim={1in 0.75in 1in 1.75in},clip]{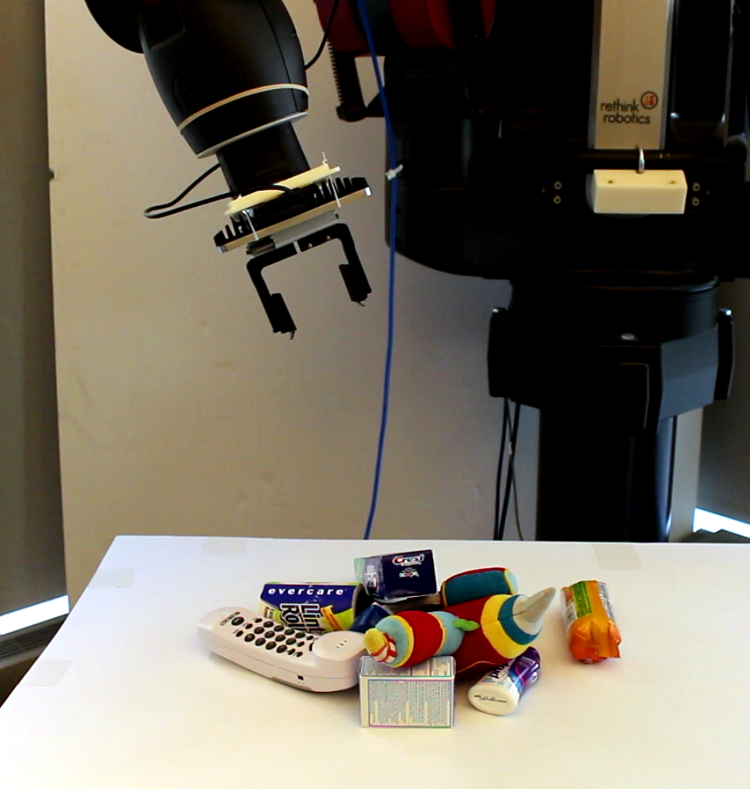}
    \includegraphics[height=1.3in,trim={1.5in 0.50in 2in 0.25in},clip]{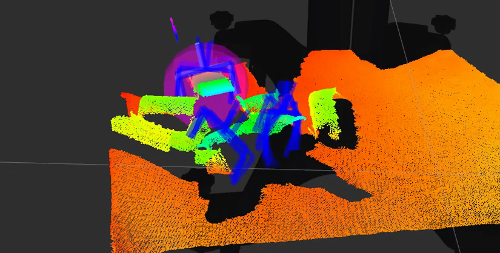}}
    \subfigure[Second view (smart)]{
    \includegraphics[height=1.3in,trim={1in 0.75in 1in 1.75in},clip]{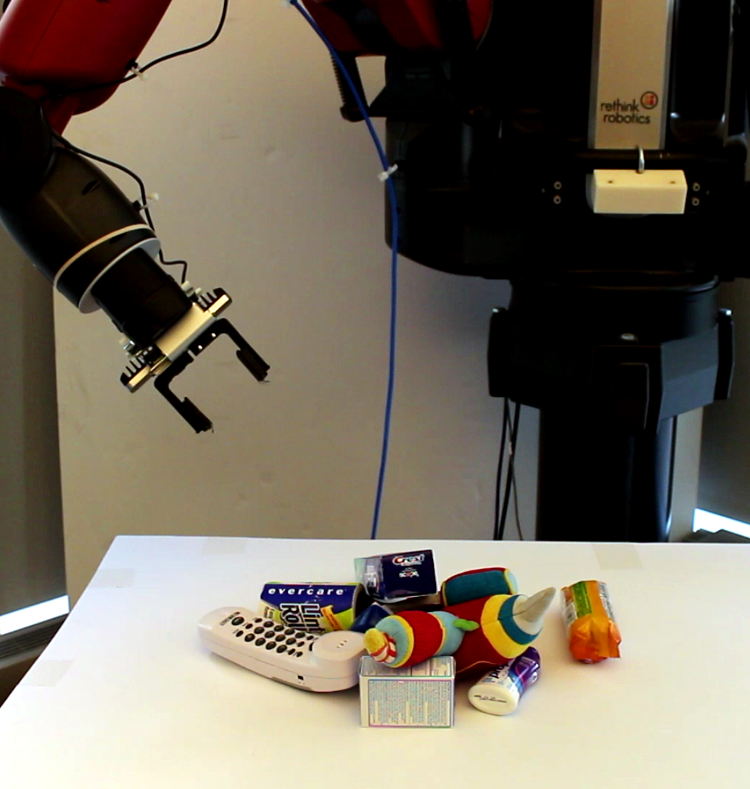}
    \includegraphics[height=1.3in,trim={1.5in 0.50in 2in 0.25in},clip]{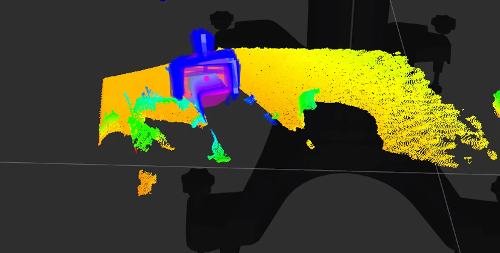}}
    \subfigure[Third view (alignment)]{
    \includegraphics[height=1.3in,trim={1in 0.75in 1in 1.75in},clip]{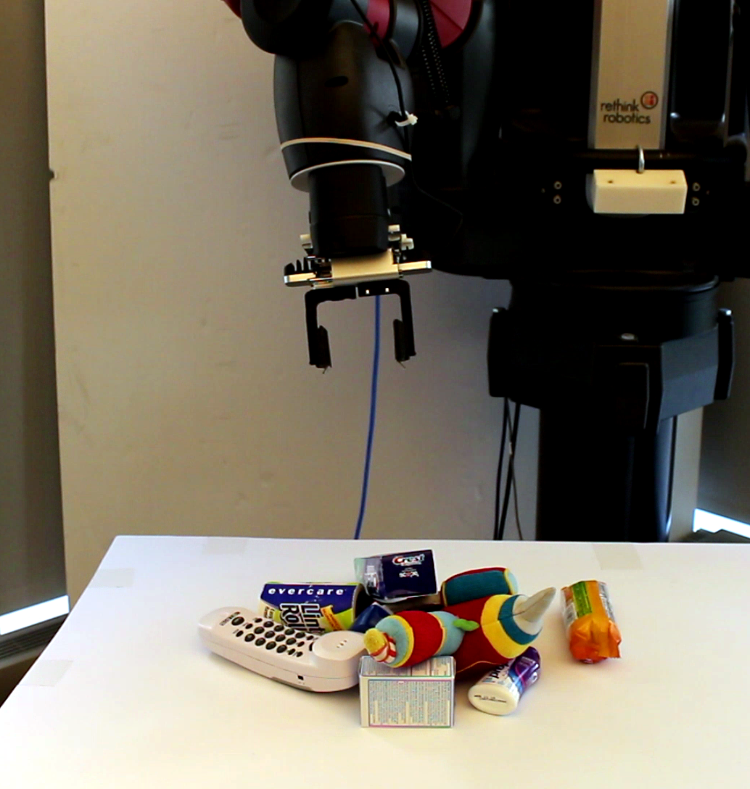}
    \includegraphics[height=1.3in,trim={1.5in 0.50in 2in 0.25in},clip]{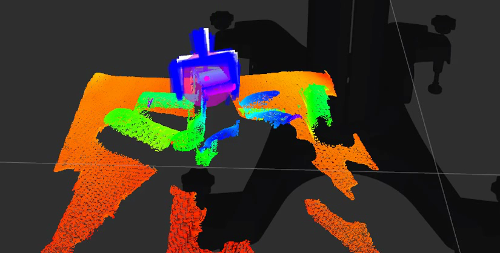}}
  \caption{Illustration of the three-view grasp detection strategy. The point cloud and grasps (right) are obtained from the depth sensor (left). The white grasp is the target grasp used for planning the next step.}
  \vspace{-0.15in}
\label{fig:sequentialView}
\end{figure}

In these experiments, each grasp proceeds as follows. First, we obtain a target grasp by taking a view of the objects from a random viewpoint and running grasp detection on this point cloud. We select one of the detected grasps based on confidence score and task-specific heuristics (such as height in the pile and how close to vertical the approach vector is). Then, we use the smart viewpoint selection strategy to obtain a viewpoint from which to view the target grasp (we select the best viewpoint subject to inverse kinematics constraints). Next, we detect grasps in this new point cloud. Since we already know the approximate location of the target grasp, we speed up the second round of grasp detection by only searching a small region (8 cm radius ball) about the grasp target. Finally, we select the highest scoring grasp (again, subject to IK constraints) and execute it. This process is illustrated in Figure~\ref{fig:sequentialView} (a-b). We measure success in terms of how often the grasp succeeds.

In principle, our proposed viewpoint selection method should use the grasp density map corresponding to the shape of the object to be grasped. However, since our scenario involves a variety of objects piled together, we just used a single map for all grasp attempts. Since the viewpoints that maximize our proposed viewpoint quality measure (density of true positives minus density of false positives) for box-like objects (Figure~\ref{fig:mapBoxesCylinders} (e)) also nearly maximizes the measure for cylinder-like objects (Figure~\ref{fig:mapBoxesCylinders} (j)), we used the map for box-like objects all the time.

In addition to the smart and random viewpoints, we also evaluated performance for a third viewpoint taken directly in front of the target grasp ($\phi=0, \theta=0$ in Figure~\ref{fig:wristSensor} (b)). We moved the sensor as close to the target grasp as possible while remaining outside the minimum viewing depth for the sensor (20 cm in our case). Instead of running the full grasp detection algorithm on the view obtained from this perspective, we just ran the candidate generation part of the algorithm and accepted the candidate most closely aligned with the target grasp. We call this the \textit{alignment} view. Its purpose is to help correct for kinematic errors in the robot: the effect of the robot's kinematic errors is limited to only those errors that accumulate while the arm travels from the the view pose to the grasp target. In order to isolate the effects of the various different views, we performed experiments for all relevant variations on viewing order: 1-2-3 (random, smart, alignment), 1-2 (random, smart), and 1-3 (random, alignment).


\begin{figure}[h]
  \centering
  \vspace{0.1in}
  \subfigure[]{\includegraphics[height=1.2in,trim={1.5in 1.5in 1.5in 0in},clip]{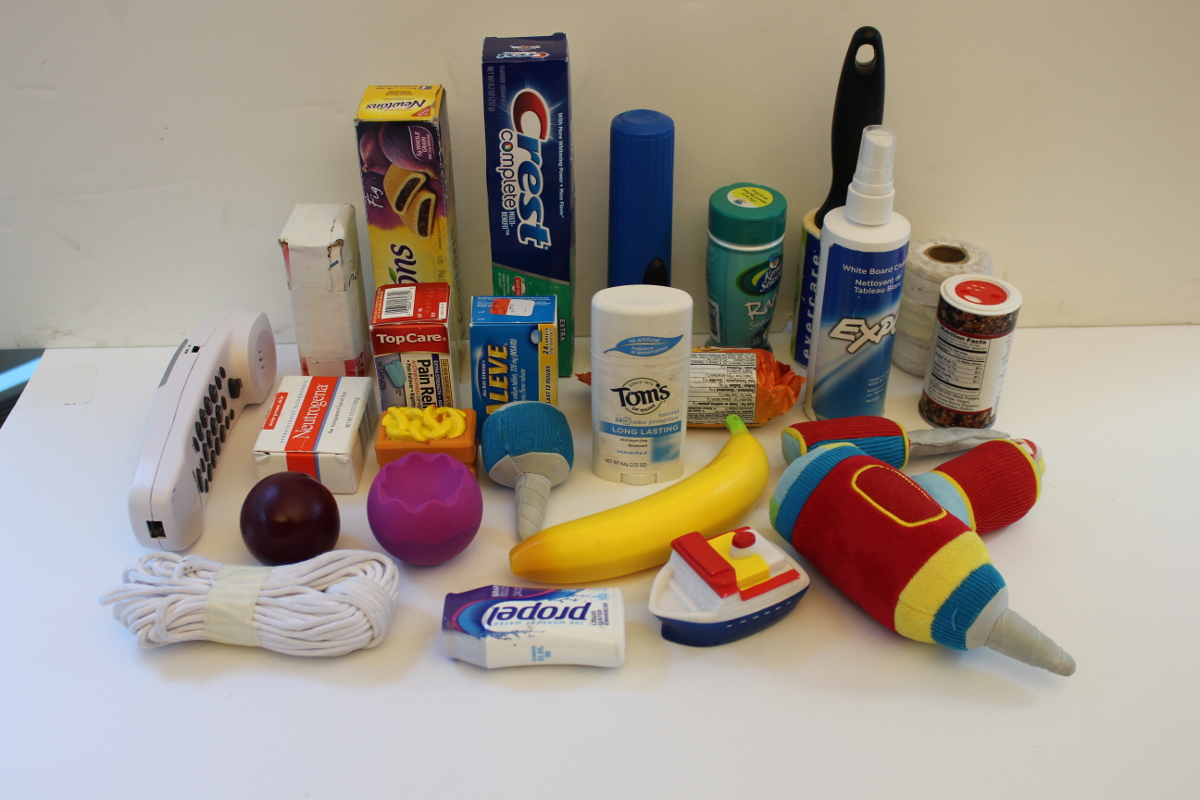}}
  \subfigure[]{\includegraphics[height=1.2in,trim={3in 0.8in 3in 0.8in},clip]{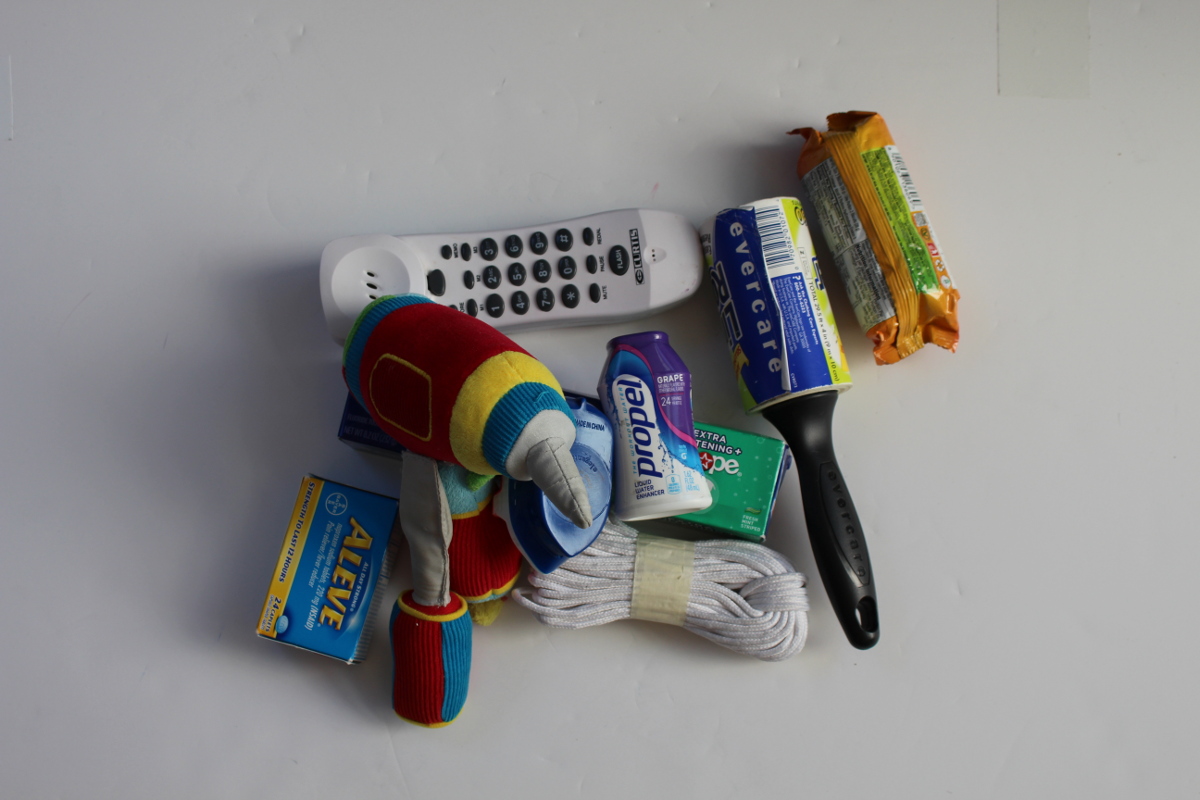}}
  \caption{(a) All 25 objects used in robot experiments. (b) Cluttered pile of 10 objects that the robot must clear.}
  \vspace{-0.15in}
\label{fig:experimentObjects}
\end{figure}

\begin{figure*}
  \centering
  \vspace{0.1in}
  \includegraphics[height=1.1in,trim={1in 1.5in 2in 10in},clip]{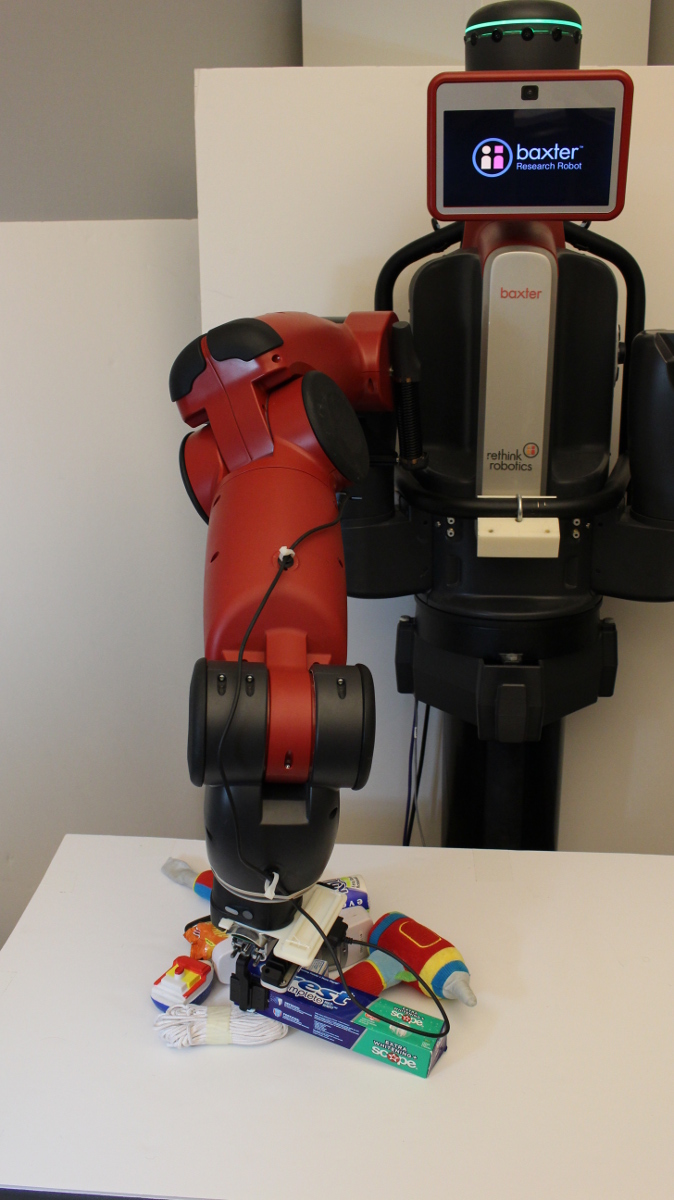}
  \includegraphics[height=1.1in,trim={1in 1.5in 2in 10in},clip]{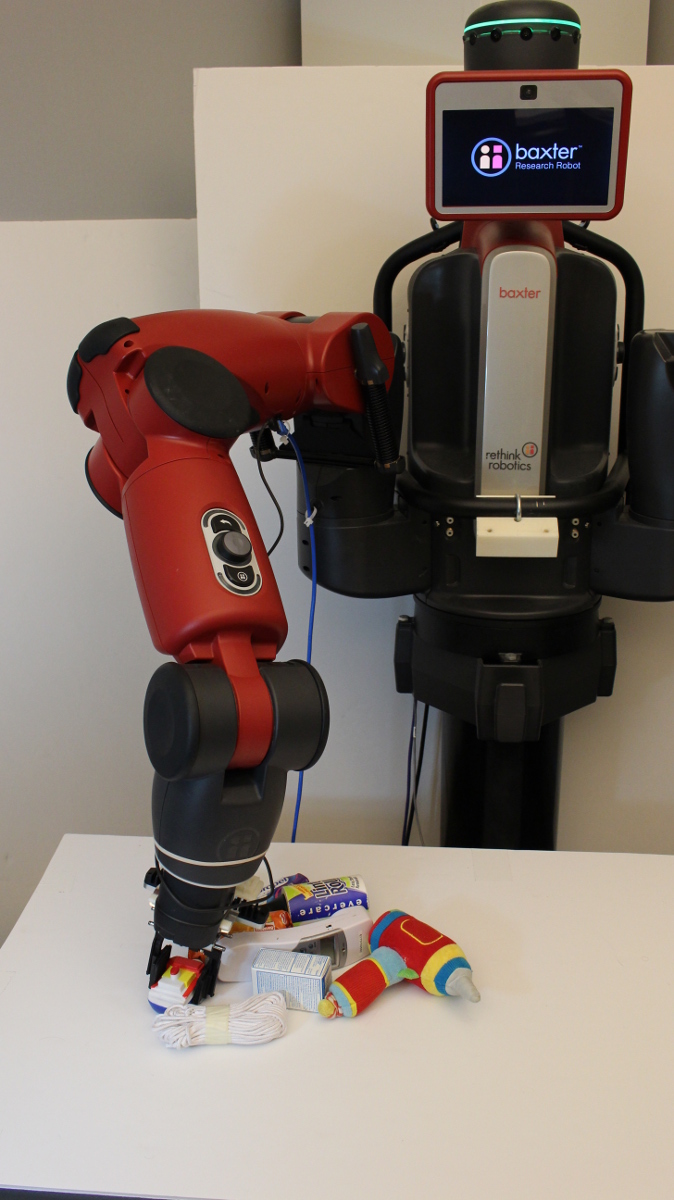}
  \includegraphics[height=1.1in,trim={1in 1.5in 2in 10in},clip]{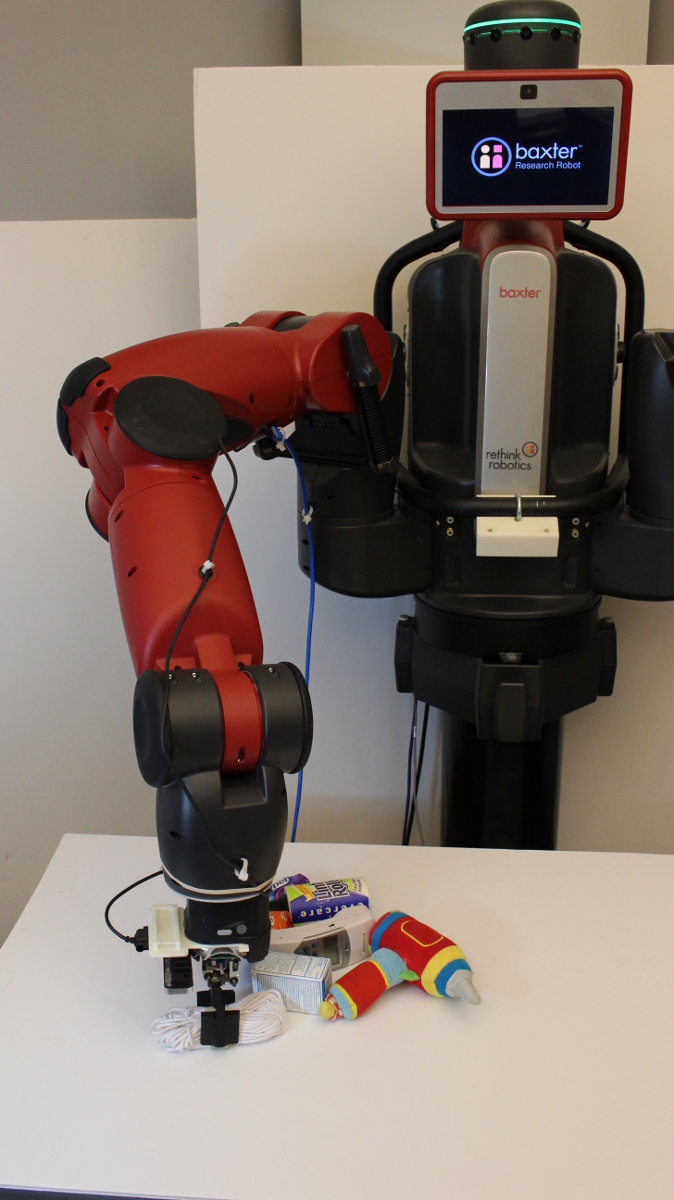}
  \includegraphics[height=1.1in,trim={1in 1.5in 2in 10in},clip]{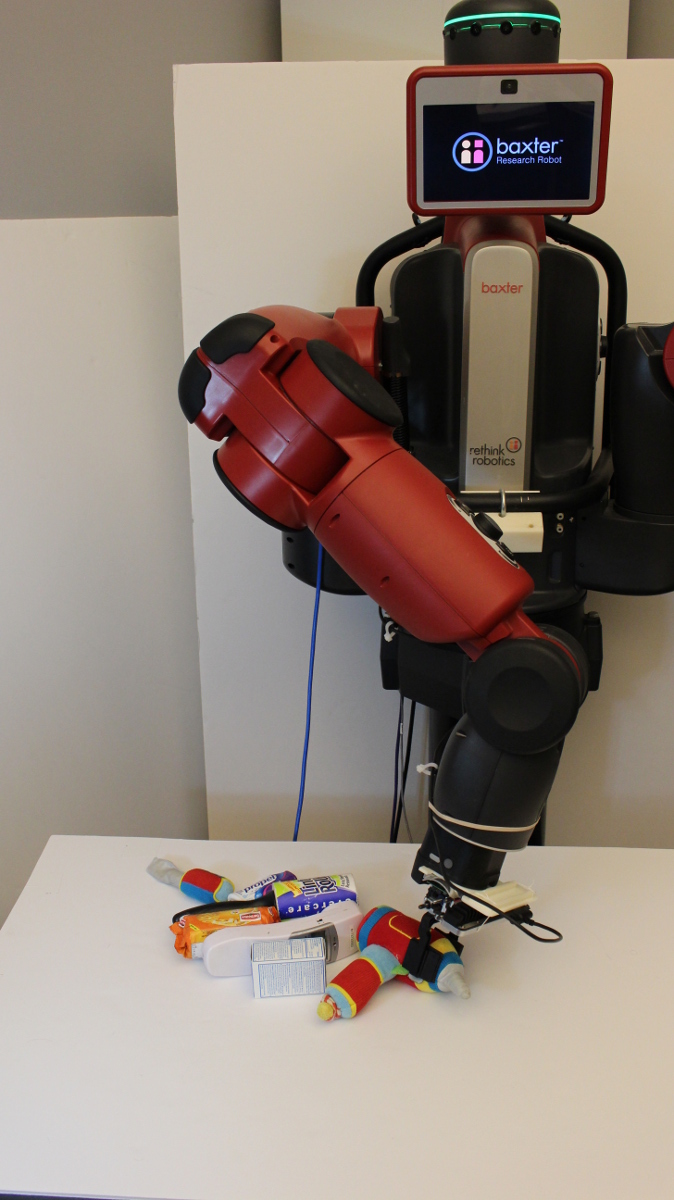}
  \includegraphics[height=1.1in,trim={1in 1.5in 2in 10in},clip]{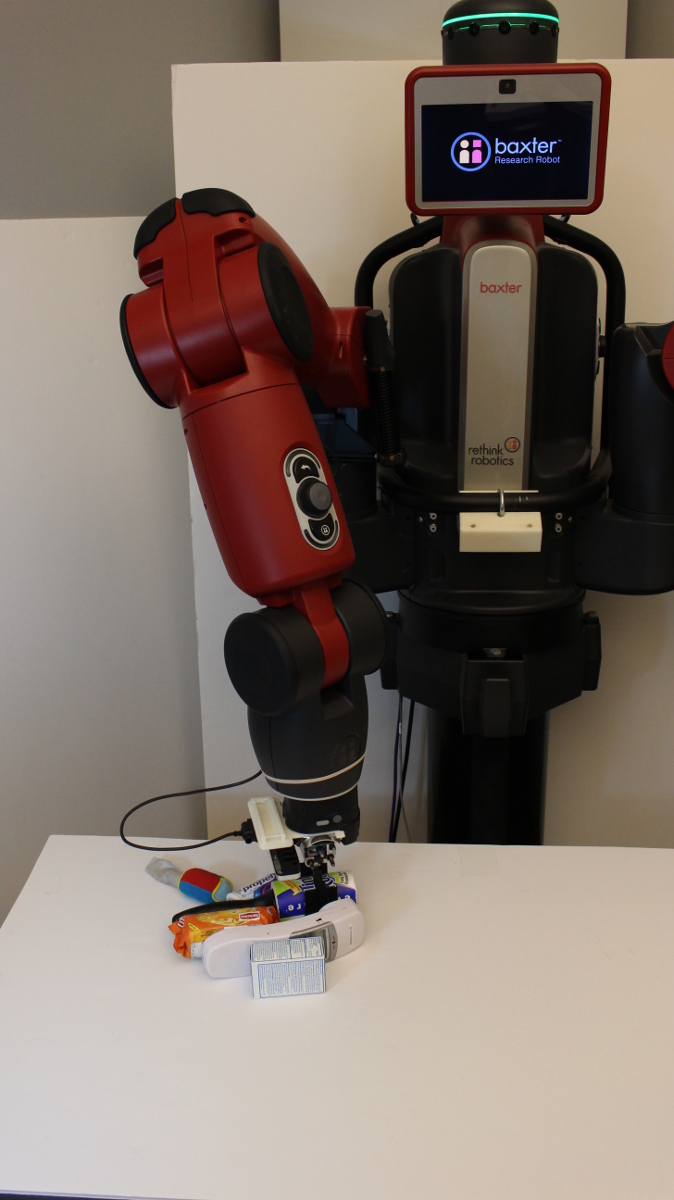}
  \caption{First 5 grasps of a typical experiment trial. All grasps were successful. This was a run with all 3 views included.}
  \vspace{-0.15in}
\label{fig:experimentGrasps}
\end{figure*}

The experimental protocol followed for each variation is similar to the one proposed in our prior work \cite{Gualtieri2016}. First, 10 objects are selected at random from a set of 25 and ``poured'' into a pile in front of the robot. (See Figure~\ref{fig:experimentObjects} (a) for the object set and Figure~\ref{fig:experimentObjects} (b) for an example pile of clutter.) Second, the robot proceeds to automatically remove the objects one-by-one as the experimenter records successes and failures. This continues until either all of the objects have been removed, the same failure occurs on the same object three times in a row, or no grasps were found after three attempts. The sensor (Intel RealSense SR300) and gripper hardware used in the experiment are shown in Figure~\ref{fig:wristSensor} (a). Figure~\ref{fig:experimentGrasps} shows the robot performing the first five grasps of one round of an experiment.~\footnote{A video illustrating the experiment is available at \url{https://youtu.be/iGRbqFsNgzo}.}

\subsection{Results}
\label{sec:robotResults}

\begin{table}[h]
\centering
\begin{tabular}{c | c | c | c | c } 
 \hline
              & Views 1-2-3 & Views 1-2 & Views 1-3 & Views 1 \\
 \hline
 Attempts     &         131 &       141 &       154 &     153 \\
 \hline
 Failures     &          17 &        31 &        26 &      39 \\
 \hline
 Success Rate &        0.87 &      0.78 &      0.83 &    0.75 \\
 \hline
\end{tabular}
\caption{Grasp success rates for the four experimental strategies.}
\vspace{-0.15in}
\label{table:robotResultsSummary}
\end{table}

Grasp success rates for the four experimental strategies are shown in Table~\ref{table:robotResultsSummary}. The ``attempts'' row of the table shows the number of grasp attempts made using each strategy, the ``failures'' row shows the number of failures for each strategy, and the ``success rate'' row is one minus the ratio between failures and attempts. The grasp failures in our experiments primarily fell into two categories as shown in Table~\ref{table:robotErrorSources}. The ``FK'' row in Table~\ref{table:robotErrorSources} denotes the number of grasp failures caused by a difference between the planned grasp and the actual grasp in the real world. The ``grasp'' row denotes the number of grasp failures caused by a detection error. The failure modes seem to reinforce intuition about what should happen if either view 2 (smart) or view 3 (alignment) is skipped. If the alignment view is skipped, then we obtain a large number of FK failures, presumably because we are not registering the point cloud close to the final grasp configuration. If the smart view is skipped, then we obtain relatively more detection failures because the algorithm does not get the best view of the target grasp (in terms of detector performance).

Perhaps the most noticeable result from this experiment is that adding the third view helps: going from 1-2 to 1-2-3 adds 9\% to the grasp success rate; going from 1 to 1-3 adds 8\%. However, this is increase (and the relatively poor performance of grasping without the alignment view) is a result of kinematic errors in the Baxter robot and/or calibration errors in the Intel RealSense SR300 sensor we used. Adding the alignment view helped to correct for these errors. However, the benefits of adding the alignment view should not overshadow the additional benefit of the smart view. Without the alignment view, adding the smart view increases the grasp success rate from $0.75$ to $0.78$ (a 3\% increase). With the alignment view, adding the smart view increases grasp success from $0.83$ to $0.87$ (a 5\% increase).

\begin{table}[h]
\centering
\begin{tabular}{c | c | c | c | c } 
 \hline
 (Error Type) & Views 1-2-3 & Views 1-2 & Views 1-3 & Views 1 \\
 \hline
 FK           &          10 &        21 &         9 &      28 \\
 \hline
 Grasp        &           4 &        10 &        14 &       9 \\
 \hline
 Other        &           3 &         0 &         3 &       2 \\
 \hline
\end{tabular}
\caption{Counts by failure type for the four experimental strategies. ``FK'' means a forward kinematics error led to the failure and ``Grasp'' means a defect in the detected grasp led to a failure.}
\vspace{-0.15in}
\label{table:robotErrorSources}
\end{table}

\section{Conclusion}

Our main conclusion is viewpoint can have a significant effect on the performance of grasp detection. The right viewpoint can enable grasp detection to find 4-6 times the number of good grasps relative to an uninformed view. Our results show this increase in the number of detected grasps can have a significant effect on the average accuracy of the top detected grasps. These results are borne out in robotic experiments on our Baxter showing an improvement in overall grasp success rates using an informed viewpoint selection method.

\FloatBarrier
\bibliographystyle{IEEEtran}
\bibliography{References.bib}

\end{document}